\PassOptionsToPackage{table,xcdraw,prologue}{xcolor} % 提前统一选项
\documentclass[10pt,twocolumn,letterpaper]{article}

\usepackage{iccv}              % To produce the CAMERA-READY version
\usepackage[accsupp]{axessibility}
\usepackage{xcolor} % 只加载一次
\usepackage{amssymb}
\usepackage{pifont}
\usepackage{amsmath}
\usepackage{amsthm}

\usepackage{booktabs}
\usepackage{makecell}
\usepackage{multirow}
\usepackage{graphicx}
\usepackage{algorithm} 
\usepackage{algpseudocode} % pseudo-code

\definecolor{iccvblue}{rgb}{0.21,0.49,0.74}
\usepackage[pagebackref,breaklinks,colorlinks,allcolors=iccvblue]{hyperref}

\definecolor{refred}{rgb}{0.21,0.49,0.74}
\definecolor{myye}{RGB}{255,249,236}
\definecolor{mygray}{RGB}{248,248,249}

\def\paperID{8394} % *** Enter the Paper ID here
\def\confName{ICCV}
\def\confYear{2025}

\title{FedDifRC: Unlocking the Potential of Text-to-Image Diffusion Models in Heterogeneous Federated Learning}

\author{Huan Wang\textsuperscript{1,*}, Haoran Li\textsuperscript{1,3,*}, Huaming Chen\textsuperscript{2}, Jun Yan\textsuperscript{1}, Jiahua Shi\textsuperscript{3}, 
Jun Shen\textsuperscript{1,\dag}\\
\textsuperscript{1}School of Computing and Information Technology, University of Wollongong, Wollongong, Australia\\
\textsuperscript{2}School of Electrical and Computer Engineering, The University of Sydney, Sydney, Australia\\
\textsuperscript{3}QLD Alliance for Agriculture and Food Innovation, The University of Queensland, Brisbane, Australia\\
{\tt\small hw226@uowmail.edu.au} \quad 
\textsuperscript{*}{\tt\small Equal Contribution} \quad 
\textsuperscript{\dag}{\tt\small Corresponding Author} 
}

\begin{document}
\maketitle
\begin{abstract}
Federated learning aims at training models collaboratively across participants while protecting privacy. However, one major challenge for this paradigm is the data heterogeneity issue, where biased data preferences across multiple clients, harming the model's convergence and performance. 
In this paper, we first introduce powerful diffusion models into the federated learning paradigm and show that diffusion representations are effective steers during federated training. To explore the possibility of using diffusion representations in handling data heterogeneity, we propose a novel diffusion-inspired \textbf{Fed}erated paradigm with \textbf{Dif}fusion \textbf{R}epresentation \textbf{C}ollaboration, termed \texttt{FedDifRC}, leveraging meaningful guidance of diffusion models to mitigate data heterogeneity. 
The key idea is to construct text-driven diffusion contrasting and noise-driven diffusion regularization, aiming to provide abundant class-related semantic information and consistent convergence signals. 
On the one hand, we exploit the conditional feedback from the diffusion model for different text prompts to build a text-driven contrastive learning strategy. On the other hand, we introduce a noise-driven consistency regularization to align local instances with diffusion denoising representations, constraining the optimization region in the feature space. 
In addition, \texttt{FedDifRC} can be extended to a self-supervised scheme without relying on any labeled data. We also provide a theoretical analysis for \texttt{FedDifRC} to ensure convergence under non-convex objectives. 
The experiments on different scenarios validate the effectiveness of \texttt{FedDifRC} and the efficiency of crucial components. Code is available at \href{https://github.com/hwang52/FedDifRC}{https://github.com/hwang52/FedDifRC}.
\end{abstract}
\section{Introduction}
\label{sec:intro}
\begin{figure}[t]
    \centering
    \includegraphics[width=1.0\linewidth]{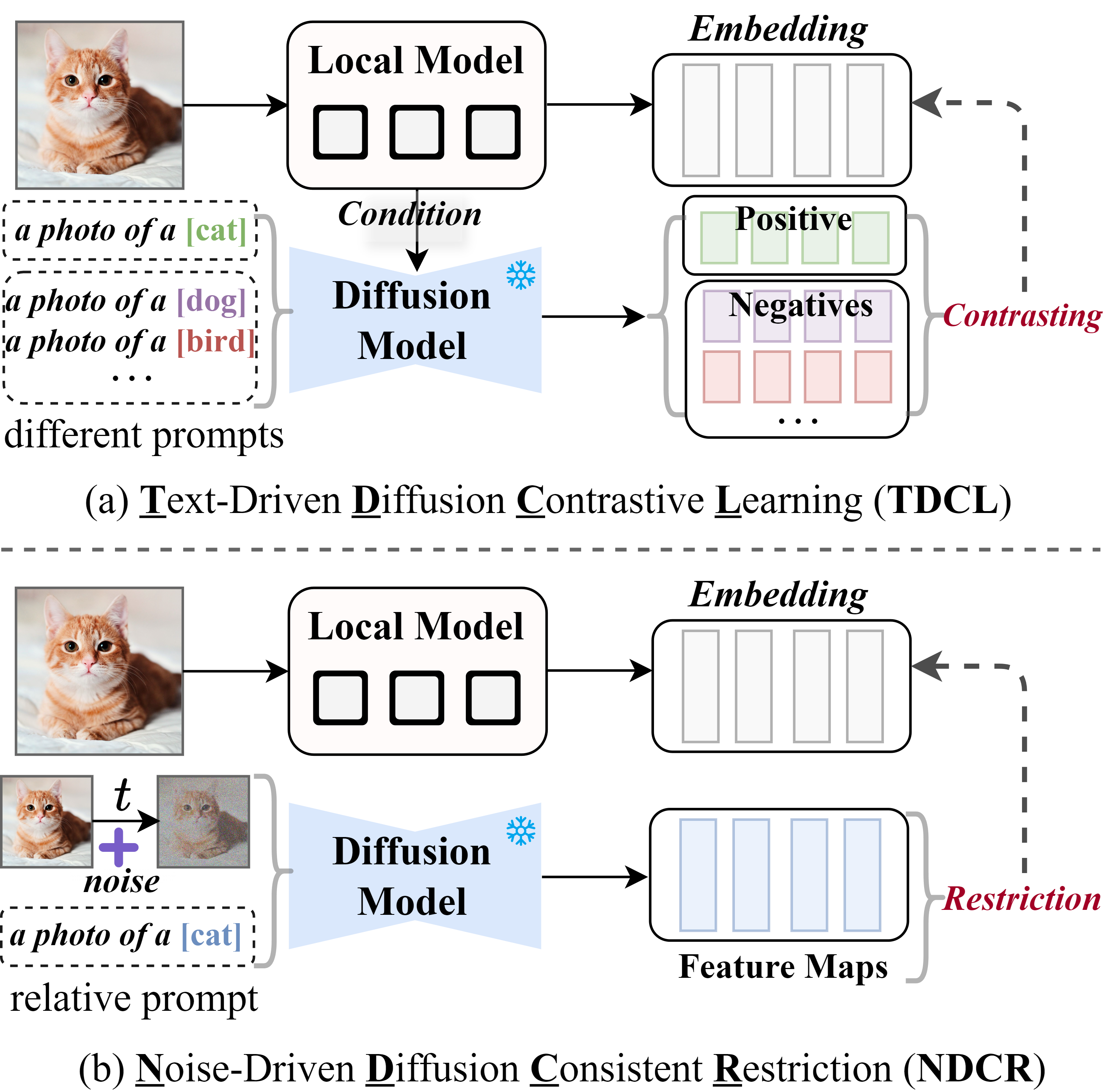}
    \caption{\textbf{The main idea of \texttt{FedDifRC} with two key modules TDCL and NDCR}. \textbf{(a)} We leverage conditional generative feedback of the pre-trained diffusion model on various text prompts to devise a \textit{text-driven contrastive learning} strategy, where the input embedding is injected as a condition. \textbf{(b)} We extract the diffusion denoising feature maps of intermediate layers from the pre-trained diffusion model at a specific time step $t$ during the diffusion backward process to construct a \textit{noise-driven regularization}.}
    % \vspace{-3mm}
    \label{fig:fig1}
\end{figure}

Federated Learning (FL)~\cite{mcmahan2017communication} has become increasingly popular as a decentralized training paradigm, with multiple distributed clients train their local models without sharing their private dataset~\cite{li2020federated,kairouz2021advances} (\textit{i.e.}, training client models locally and aggregating them globally~\cite{li2020fedprox}). 
One major challenge faced in FL is the potential discrepancies in the local data distributions across distributed clients, which is known as the data heterogeneity issue~\cite{mendieta2022local,qu2022rethinking,luo2021no}. 
Specifically, the private data is collected from different sources with diverse preferences and presents non-independent and identically distributed (\textit{non-iid}) distribution~\cite{zhu2021federated,li2022federated}. 
Such \textit{non-iid} data discrepancies can lead to inconsistency in the clients' local objective functions and optimization directions, making the model convergence slow and unstable~\cite{karimireddy2020scaffold,li2020fedprox,Li2020On,pathak2020fedsplit,yan2024federated}.
% In addition, the aggregated global model lacks a convergence guarantee, which unavoidably faces compromised convergence speed and confined performance improvement~\cite{Li2020On,pathak2020fedsplit,yan2024federated}.

To mitigate the negative impact of the data heterogeneity, a mainstream of subsequent efforts delves into introducing a variety of suited signals to regulate the private model, there are mainly two perspectives: \romannumeral1) local model optimization at the client side and \romannumeral2) global model aggregation at the server side. 
The former focuses on optimization strategies to make the variance of local updates limited~\cite{li2020fedprox,gao2022feddc}. Another aims to improve the efficacy of global model aggregation~\cite{ye2023feddisco,zhang2023fedala,nguyen2022federated}. 
In fact, most of these previous methods force the local models to be consistent to the global model. Although they have a certain effect by reducing gradient inconsistency, the gradually enlarged parameter deviation persists. Hence, we argue that local model overfits the local data distribution, which is magnified under the data heterogeneity and thus hinders the improvement of model's performance in FL.
% (\textit{e.g.}, label shift, and feature shift). 
%Hence, we argue that naive learning on local private data brings poor generalizable ability: \textit{local private model overfits local domain distribution}, which is magnified under the data heterogeneity and thus hinders the improvement of generalizability and discriminability.

Recently, the development of diffusion models (DMs)~\cite{ho2020denoising,rombach2022high,nichol2021improved,croitoru2023diffusion} provides fresh opportunities. The diffusion models have demonstrated remarkable power in generating the diverse and realistic images, showing strong ability to capture intricate details and patterns from underlying data. For instance, with proper guidance, stable diffusion (SD)~\cite{rombach2022high} models can effectively align a text prompt with the corresponding image context, which suggests their potential ability to understand both high-level and low-level contextual visual concepts regarding what an image contains. This semantic generative and visual comprehension capabilities of SD models motivate us to think: \textit{is it possible to exploit the inherent fine-grained visual representation capabilities of diffusion models to promote heterogeneous FL training?}

Taking into account both the effectiveness and efficiency, we revisit the underlying representations in the SD models, which are intermediate feature maps in a diffusion process (\textit{e.g.}, denoise or noise an image) produced by multiple calling a UNet~\cite{ronneberger2015u}. We argue that diffusion representations are effective steers for FL training: \romannumeral1) extensive general knowledge inherent in diffusion models enhance local semantic diversity in FL; \romannumeral2) smooth correspondences about semantic objects in diffusion models is a natural guiding signal in FL. Building on these two insights, we first propose \textbf{\textit{conditional diffusion representations}} by imposing various text prompts in the SD model. In this way, each instance is contrastively enhanced with a set of text-driven diffusion representations, capturing rich domain variance. Then, we devise \textbf{\textit{denoising diffusion representations}} from semantic relations between visual and linguistic concepts in the SD model. We extract the feature maps from different hierarchies of the UNet decoder in the SD model's denoising process, serving as guidance signals to avoid over-optimization in FL local training.% We will analyze the superiority in Sec.~\ref{sec:sec33} and Sec.~\ref{sec:sec34}.

In this paper, we propose a diffusion-inspired \textbf{Fed}erated framework with \textbf{Dif}fusion \textbf{R}epresentation \textbf{C}ollaboration, as \texttt{FedDifRC}, to handle the data heterogeneity issue in FL, which consists of two key modules (see Fig.~\ref{fig:fig1}): \romannumeral1) \textbf{First}, we devise Text-driven Diffusion Contrastive Learning (TDCL), which utilizes conditional diffusion representations to construct inter-contrastive learning~\cite{chen2020simple,he2020momentum,wang2021understanding,li2021model} in a text-driven diffusion manner (the sample's embedding is injected as a condition). By maximizing the agreement between the local instance and its corresponding representations, TDCL adaptively captures class-relevant information and semantically meaningful knowledge. This objective encourages sample's embedding closer to conditional diffusion representations with same semantics and away from other negatives, thus maintaining a clear decision boundary. \romannumeral2) \textbf{Second}, we use the denoising diffusion representations to build a flat and stable convergent target and further propose Noise-driven Diffusion Consistent Restriction (NDCR). Concretely, we extract denoising feature maps of the noisy input sample in the SD model's backward process, as a regularization term to constrain the optimization region of the local model. The local sample is forced to align the embedding with its denoising representations at the feature space to promote the uniformity for local training.

%In addition, \texttt{FedDifRC} can be extended to a self-supervised scheme without relying on any labeled data (Sec.~\ref{sec:ssl}). We also provide a convergence guarantee for \texttt{FedDifRC} under non-convex case (Sec.~\ref{sec:the}). 
Our main contributions are summarized as follows:
\begin{itemize}
    \item We explore the possibility of using diffusion models to tackle data heterogeneity in FL and devise the conditional and denoising diffusion representations as effective steers to regulate model training in heterogeneous FL.
    \item We propose \texttt{FedDifRC}, a diffusion-inspired FL framework to handle data heterogeneity by leveraging the complementary advantages of two key components: \romannumeral1) TDCL facilitates the local model to learn more general class-relevant knowledge; \romannumeral2) NDCR enforces the uniformity in the feature space for local training.
    \item We conduct extensive experiments for different FL scenarios (\textit{e.g.}, label shift, domain shift, and long-tailed) and different application scenarios (\ie, intra-domain and cross-domain) to demonstrate the effectiveness of our \texttt{FedDifRC} and the indispensability of each component.
\end{itemize}
\section{Related Work}
\label{sec:related}

\subsection{Heterogeneous Federated Learning}
Federated learning is proposed to address privacy concerns in a distributed learning environment (\textit{e.g.}, Federated Averaging (FedAvg)~\cite{mcmahan2017communication}). Several studies have improved the local update process on the client side or the global aggregation process on the server side to tackle data heterogeneity in FL. 
The former focuses on local optimization strategies to make the diversity between the client model and the global model limited in the parameter level~\cite{li2020fedprox,karimireddy2020scaffold,gao2022feddc,mendieta2022local,huang2022learn,tan2022fedproto,li2021model,huang2023rethinking}. 
The another branch aims to adopt aggregation mechanisms on the server to alleviate the negative influence of data heterogeneity~\cite{qi2024model,ye2023feddisco,zhang2023fedala,wang2023fedcda,lee2023layer,wang2024aggregation}. 
Recently, existing FL works that apply diffusion models mainly focus on building synthetic datasets~\cite{yang2024exploring,li2024feddiff,zhao2023federated,stanley2024phoenix}. With the assistance of synthetic datasets, the FL global model can be retrained by this data augmentation scheme to mitigate data heterogeneity. Although these generated data have comparable quality and diversity to client images, the local model may overfit local domain distribution, and the inherent heterogeneity across clients may not have been addressed. In contrast, we focus on how to effectively leverage the fine-grained visual representation from diffusion models to promote a generalizable FL global model.

% In contrast, in this paper, we focus on \textit{how to effectively leverage meaningful internal knowledge from the pre-trained diffusion process, to pave the way for applying diffusion representations on federated learning, to promote a generalizable FL global model}.

\subsection{Diffusion Model Representations}
Recently, diffusion models have shown significant advances in image generation, editing, and stylized~\cite{ho2020denoising,rombach2022high,nichol2021improved,croitoru2023diffusion,ling2023diffusion}, suggesting that they contain rich underlying representations to be exploited for different downstream tasks. 
DIFT~\cite{tang2023emergent} is used to extract the diffusion features without explicit supervision. Diffusion Hyperfeatures~\cite{luo2024diffusion} integrates multi-scale and multi-timesteps diffusion features into feature descriptors. 
The diffusion representations have shown impressive results on the vision branch, including being testified as certifiable classifiers~\cite{li2023your,clark2024text,wei2023diffusion} and semantics correspondence learners~\cite{tang2023emergent,zhang2024tale,xiang2023denoising}. 
In this paper, we expand the diffusion models to FL and construct the conditional and denoising diffusion representations, offering fruitful class-relevant information and consistent convergence signals.

% ControlNet~\cite{zhang2023adding} is used to add conditioning controls to large diffusion models, which preserves the quality and capabilities of the diffusion model by locking its parameters. 
% Diffusion representations have shown impressive results on the vision branch, including being testified as certifiable classifiers~\cite{li2023your,clark2024text} and semantics correspondence learners~\cite{tang2023emergent,zhang2024tale}. 
% In this paper, \textit{we further expand to the FL and construct conditional and denoising diffusion representations as effective steers to regulate local training, offering fruitful class-relevant semantics information and consistent convergence signals for federated training}.

\subsection{Contrastive Learning}
Contrastive learning (CL) has been widely applied to self-supervised learning scenarios~\cite{gui2024survey,chen2020simple,he2020momentum}. Many works are focused on learning the encoder where the embeddings of the same sample are pulled closer and those of different samples are pushed apart~\cite{tian2020makes,chuang2020debiased,cui2021parametric,khosla2020supervised}. They mainly construct a positive pair and some negative pairs for each data instance and use InfoNCE~\cite{oord2018representation} to contrast positiveness against negativeness. There are some works that incorporate contrastive learning into FL to assist local training to achieve higher model performance~\cite{li2021model,huang2022learn,han2022fedx,tan2022fedpcl}. 
In our work, Text-driven Diffusion Contrastive Learning (TDCL) is devised to promote local model by attracting samples to positive representations while pushing away other negatives.
% For example, SimCLR~\cite{chen2020simple} proposed a simple and effective method for the contrastive learning of visual representations by constructing positive and negative sample pairs via data augmentation. MoCo~\cite{he2020momentum} further built a dynamic dictionary with a queue and a moving-averaged encoder, via a look-up strategy to promote contrastive unsupervised learning tasks. There are some works that incorporate contrastive learning into FL to assist local training to achieve higher model performance~\cite{li2021model,huang2022learn,han2022fedx}, focusing on building auxiliary datasets and learning shared representation. 
% Differently, in our work, Text-driven Diffusion Contrastive Learning (TDCL) is devised for promoting local training by \textit{attracting each sample to corresponding positive conditional diffusion representations, while pushing away other negative representations}.
\section{Diffusion Models Help Federated Learning}
\label{sec:meet}

In this section, we begin with a brief overview of the general FL paradigm and the generative diffusion models (Sec.~\ref{sec:sec31}). 
Then, we analyze the effectiveness of diffusion representations and illustrate our core insights (Sec.~\ref{sec:sec32}). 
Finally, we will introduce how to extract the conditional and denoising diffusion representations in Sec.~\ref{sec:sec33} and Sec.~\ref{sec:sec34}.

\subsection{Preliminaries and Background}
\label{sec:sec31}

\textbf{Federated Learning:} We focus on the case with $K$ clients holding the heterogeneous data partition $\{\mathcal{D}_{1}, \mathcal{D}_{2}, \ldots, \mathcal{D}_{K}\}$ with private dataset $\mathcal{D}_{k}=\left\{x_{i}, y_{i}\right\}_{i=1}^{n_{k}}, y_{i}\in\{1,\ldots,|\mathcal{C}|\}$ for the $k$-th client, where $n_{k}$ denotes the local dataset scale. 
Let $n_{k}^{j}$ be the number of samples with label $j$ at client $k$, and $\mathcal{D}_{k}^{j}=\{(x, y) \in \mathcal{D}_{k} | y=j\}$ denotes the set of samples with label $j$ on the $k$-th client, so the number of samples in $j$-th class is $n^{j}=\sum_{k=1}^{K}n_{k}^{j}$. 
The objective of the general FL is to learn an optimal global model $w$ across $K$ clients:
\begin{equation} \label{eq:eq1}
    \min_{w} \mathcal{L}(w) = \frac{1}{K} \sum_{k=1}^{K} \frac{n_{k}}{N} \mathcal{L}_{k} (w; \mathcal{D}_{k}),
\end{equation}
where $\mathcal{L}_{k}$ is the $k$-th client local loss function and $w$ is the global model parameters. $N$ is the total number of samples among all clients. In the FL training setup, we consider the local model $f$ with parameters $w=\left\{u, v\right\}$. It has two modules: \romannumeral1) a feature extractor $h(u)$ with the parameters $u$ maps each sample $x_{i}$ to a $d$-dim (default as 512) vector as $z_{i}=h(u;x_{i})$; \romannumeral2) a classifier $g(v)$ with parameters $v$ maps $z_{i}$ into a $|\mathcal{C}|$-dim output logits as $l_{i}=g(v;z_{i})$.

\noindent \textbf{Diffusion Models:} 
Given the image sample $x$ from an underlying distribution $p(x)$, a forward diffusion process as a Markov chain to add random Gaussian noise $\epsilon \sim \mathcal{N}(0,\mathbf{I})$ to the original sample $x_{0}$, and $x_{t}$ is the noise sample after adding $t$-steps noise:
\begin{equation} \label{eq:eq2}
    x_{t} = \sqrt{1-\gamma_{t}} x_{t-1} + \sqrt{\gamma_{t}} \epsilon_{t}, \; t \in \{1,...,T\}, 
\end{equation}
where $T$ is the number of diffusion timesteps, $\gamma_{t} \in (0,1)$ is a adjustable time variance schedule~\cite{ho2020denoising}. By leveraging the additive property of the Gaussian distribution, Eq.~\eqref{eq:eq2} can be further reformulated as $x_{t} = \sqrt{\bar{\alpha}_{t}} x_{0} + \sqrt{1-\bar{\alpha}_{t}} \epsilon_{t}$, where $\alpha_{t}=1-\gamma_{t}$ and $\bar{\alpha}_{t} \doteq \prod_{i=1}^{t} \alpha_{i}$. 
On this basis, the image $x_{0}$ can be obtained from a random noise $x_{t} \sim \mathcal{N}(0,\mathbf{I})$ by reversing the above forward diffusion process~\cite{ho2020denoising}:
\begin{equation} \label{eq:eq3}
    x_{t-1} = \frac{1}{\sqrt{\alpha_{t}}} (x_{t}-\frac{1-\alpha_{t}}{\sqrt{1-\bar{\alpha}_{t}}}\epsilon_{\theta}(x_{t}, t)), \; t \in \{T,...,1\},
\end{equation}
where $\epsilon_{\theta}$ denotes the denoiser network with the parameters $\theta$ to predict noise $\epsilon$, and the denoiser network is usually implemented as a UNet architecture~\cite{ronneberger2015u}. 
Further, the denoiser $\epsilon_{\theta}$ is optimized using the loss $\mathcal{T}_{\theta}$ as follows:
\begin{equation} \label{eq:eq4}
   \mathcal{T}_{\theta}=\mathbb{E}_{\epsilon,x_{0},t}[|| \epsilon_{\theta}(x_{t},t)-\epsilon ||^{2}_{2}], \; \epsilon \sim \mathcal{N}(0,\mathbf{I}),
\end{equation}
where $\epsilon$ is the noise in the forward diffusion process, and $\epsilon_{\theta}$ is usually implemented as a UNet~\cite{ronneberger2015u}. 
Besides, the diffusion model's objective in Eq.~\eqref{eq:eq4} can be easily extended to a conditional diffusion paradigm by adding a condition $\mathbf{c}$ to the $\epsilon_{\theta}$. Thus, the training objective should be modified as:
\begin{equation} \label{eq:eq5}
   \mathcal{T}_{\theta}=\mathbb{E}_{\epsilon,x_{0},t,\mathbf{c}}[|| \epsilon_{\theta}(x_{t},t,\mathbf{c})-\epsilon ||^{2}_{2}], \; \epsilon \sim \mathcal{N}(0,\mathbf{I}),
\end{equation}
with the denoiser $\epsilon_{\theta}(x_{t},t,\mathbf{c})$ in Eq.~\eqref{eq:eq5}, 
we can generate data with a specific condition $\mathbf{c}$ by solving a reverse process.

\subsection{Motivation and Analysis} \label{sec:sec32}
The diffusion model is a probabilistic generative model designed for denoising by systematically reversing a progressive noising process, and recent studies~\cite{yang2023diffusion,wang2023infodiffusion} have verified that diffusion-based generative models can be effective representation learners. Stable diffusion models are able to weave from scratch with fruitful and innovative depictions of high fidelity and further capture the underlying properties, positions, and correspondences from visual concepts. Such a phenomenon shows that SD models can align a textual prompt with its corresponding image context, demonstrating their potential ability in understanding both high-level (\textit{e.g.}, semantic relations) and low-level (\textit{e.g.}, textures, edge, and structures) visual concepts in an image.

Autoencoders~\cite{zhuang2015supervised,zhang2018network} can learn to reconstruct their input at the output and have been one of the dominant approaches for representation learning. We assume a linear autoencoder expressed as $W_{a} W_{b} \cdot x + W_{c} \cdot x$ to imitate UNet~\cite{ronneberger2015u} used in diffusion models, where $W_{a} \in \mathbb{R}^{d \times d^{*}}$, $W_{b} \in \mathbb{R}^{d^{*} \times d}$, and $W_{c} \in \mathbb{R}^{d \times d}$, $d^{*} < d$. For $x_{t} = \sqrt{\bar{\alpha}_{t}} x_{0} + \sqrt{1-\bar{\alpha}_{t}} \epsilon_{t}$ and we set $\mathbb{E}[x_{0}]=0$, thus minimizing the loss $\mathcal{T}_{\theta}$ in Eq.~\eqref{eq:eq4} is equivalent to minimize the modified objective $\mathcal{T}_{\theta}^{*}$:
\begin{equation} \label{eq:eq6}
\begin{aligned}
   \mathcal{T}_{\theta}^{*} & = \mathbb{E}_{\epsilon,x_{0},t}[|| \epsilon_{\theta} (\sqrt{\bar{\alpha}_{t}} x_{0} + \sqrt{1-\bar{\alpha}_{t}} \epsilon_{t})-\epsilon ||^{2}_{2}], \\
   & \equiv \min \{ \underline{\bar{\alpha}_{t} (W_{a}W_{b}+W_{c})^{\top} \Sigma_{x_{0}} (W_{a}W_{b}+W_{c})} \},
\end{aligned}
\end{equation}
where $\Sigma_{x_{0}}$ means covariance matrix of $x_{0}$. Eq.~\eqref{eq:eq6} ensures a meaningful latent space for the diffusion models, similar to the linear autoencoder that encodes data into the principal components space. Therefore, the diffusion training process is equivalent to encourage the disentanglement of the latent representations until good-quality reconstruction is reached.

\begin{figure}[tbp]
    \centering
    \includegraphics[width=1.0\linewidth]{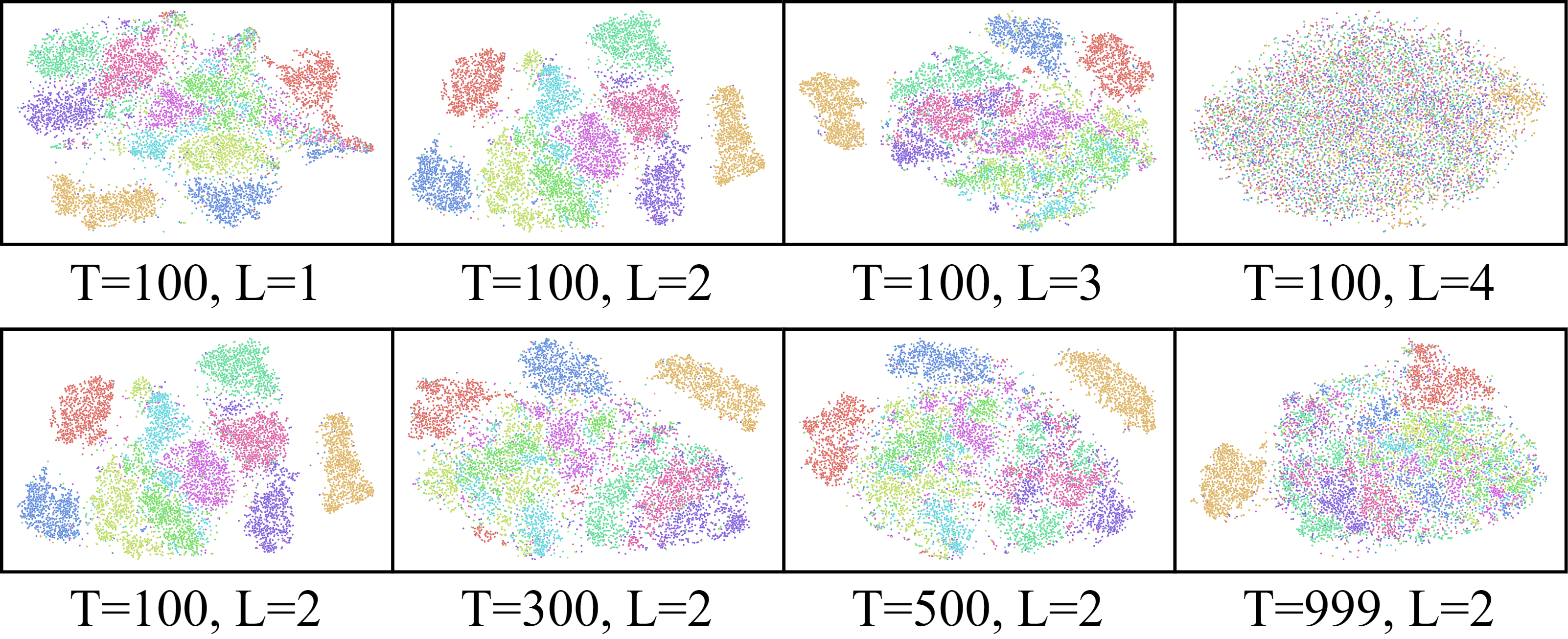}
    \caption{The t-SNE feature visualization on the MNIST dataset from the diffusion UNet decoder, \textit{without training}, where diffusion time-steps $T \in \{100,300,500,999\}$ and layer $L \in \{1,2,3,4\}$.}
    \label{fig:fig2}
\end{figure}

\begin{figure}[tbp]
    \centering
    \includegraphics[width=1.0\linewidth]{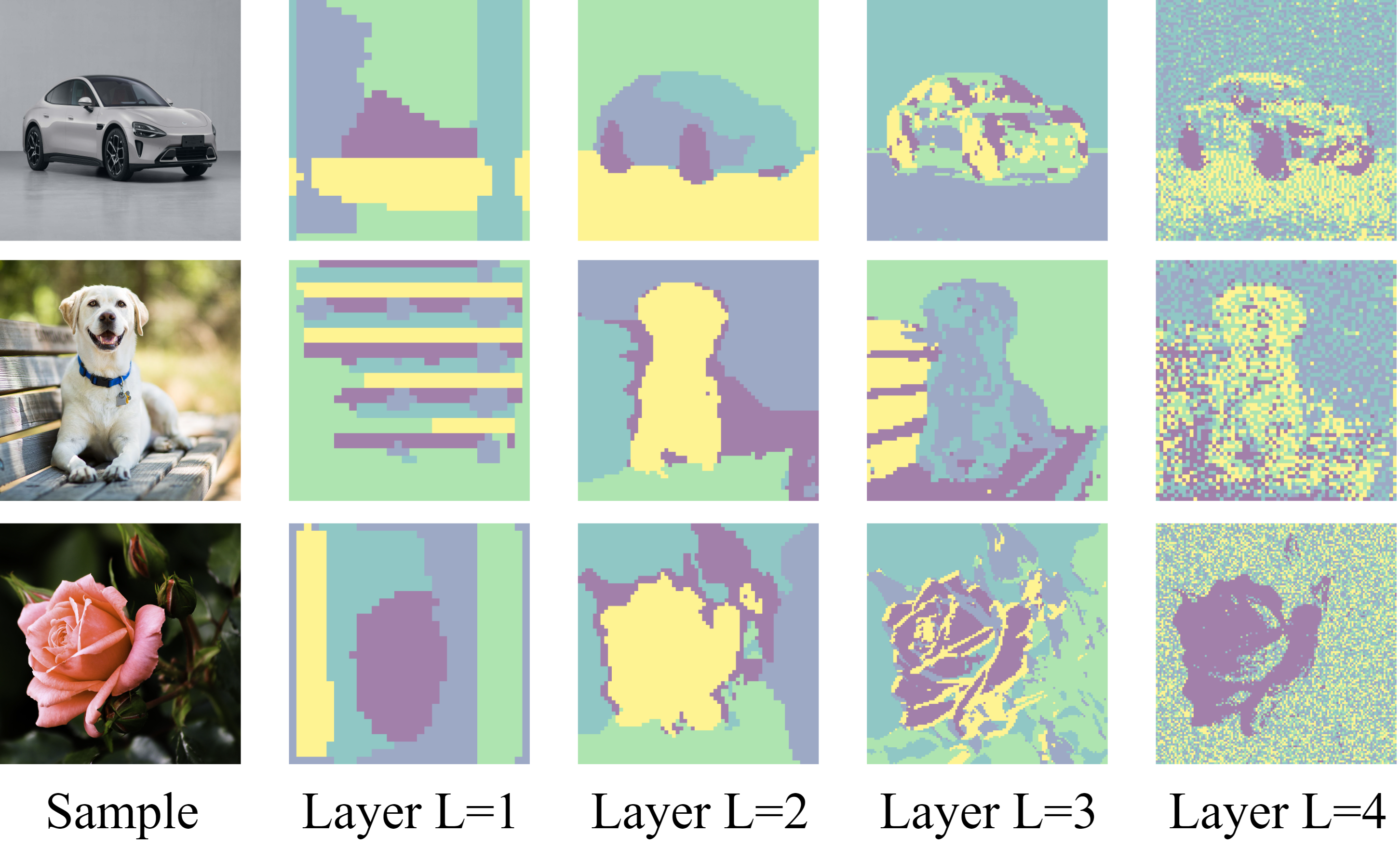}
    \caption{The K-Means clustering feature visualization (each color represents a cluster) from the diffusion UNet decoder's layers.}
    \label{fig:fig3}
\end{figure}

To validate the representation ability in diffusion models, we use t-SNE~\cite{van2008visualizing} to visualize the latent features, shown in Fig.~\ref{fig:fig2}, without training on the specific MNIST~\cite{lecun1998gradient} dataset. Concretely, we use the pre-trained SD model to extract features of the UNet decoder $\epsilon_{\theta}$, $T$ means diffusion time-steps, $L$ means the layer of the UNet decoder $\epsilon_{\theta}$ ($L \in \{1,2,3,4\}$). 
Our results in Fig.~\ref{fig:fig2} indicate that, \textit{even without supervision, diffusion models are able to group input samples} at properly selected time-steps $T$ and layer $L$. However, when the $T$ or $L$ is too large (\textit{e.g.}, $T=999$ or $L=4$), the representations become inseparable, leading to blurred decision boundaries. 
Furthermore, we apply K-Means clustering (cluster size set as 5) on the latent features from the diffusion UNet decoder and visually examine \textit{whether the diffusion features contain consistent semantic information}. 
In Fig.~\ref{fig:fig3}, different parts of the visual objects are clustered, showing that earlier layers (\textit{e.g.}, $L=2$) capture coarse yet consistent semantics, while later layers (\textit{e.g.}, $L=3$) focus on low-level textural details. 
\textit{Both t-SNE and K-Means observation results motivate us to leverage the fine-grained visual representation capabilities from the SD model to promote heterogeneous FL training}.

\begin{figure}[tbp]
    \centering
    \includegraphics[width=1.0\linewidth]{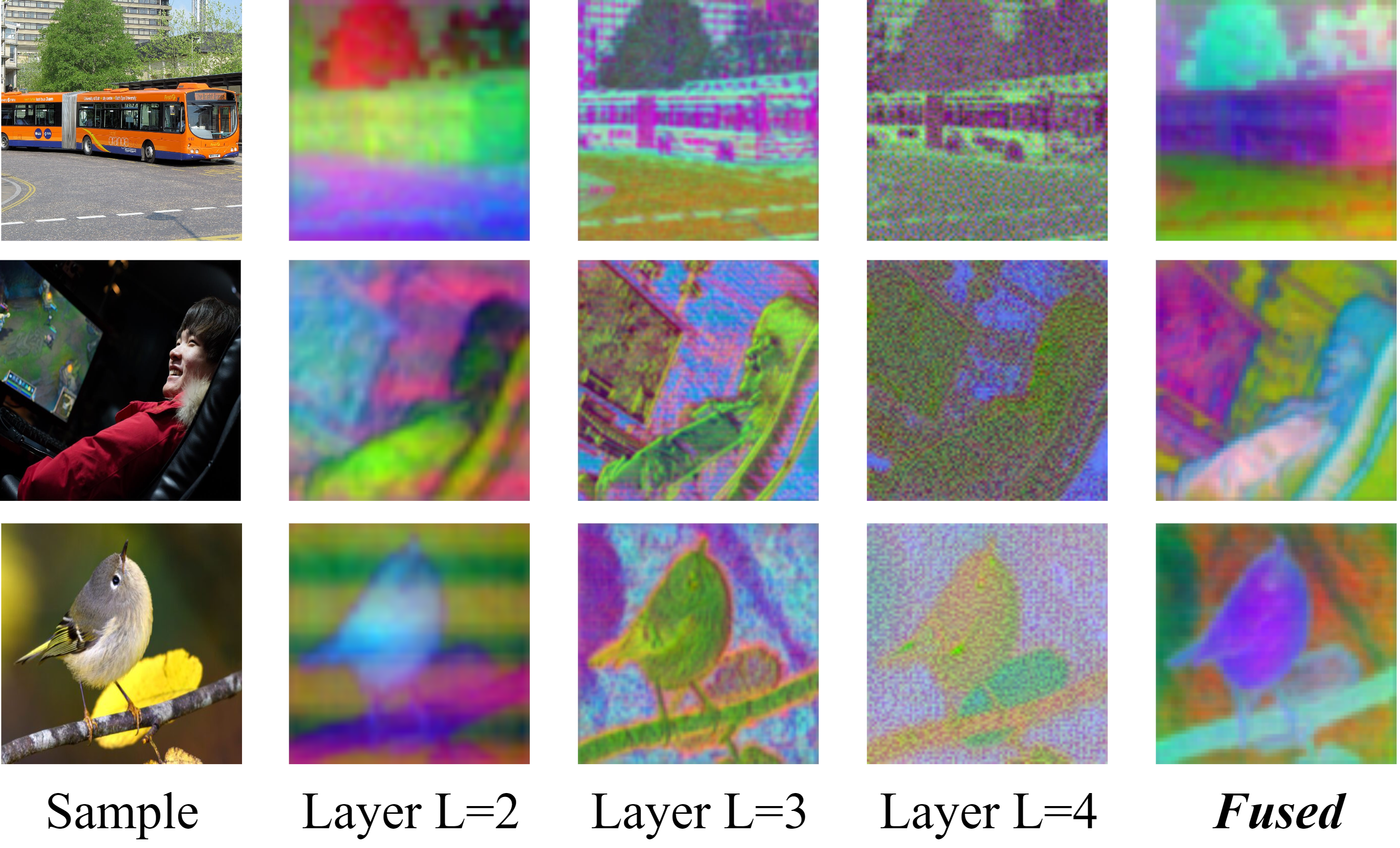}
    \caption{The PCA visualization of early (layer 2), later (layer 4), and \textit{fused features} (see Eq.~\eqref{eq:eq8}), where the first three components of PCA-computed serve as color channels for each image.}
    \label{fig:fig4}
\end{figure}

\subsection{Conditional Diffusion Representations} \label{sec:sec33}
The key idea of our conditional diffusion representations is based on cognitive diversity of generative diffusion models for different text prompts, via a condition-guided manner. 
Specifically, for the client $k$, given an input image $x_{i}$ of the local dataset $\mathcal{D}_{k}=\{x_{i}, y_{i}\}_{i=1}^{n_{k}}, y_{i} \in \{1, ..., |\mathcal{C}|\}$ means the label of $x_{i}$, and we let $\mathcal{S}$ represents the set of all class names, so we use $\mathcal{S}_{y_{i}}$ to denote the class name of the label $y_{i}$. 
From our implementation, the image $x_{i}$ is fed into the encoder $h_{k}$ of the local model $f_{k}$ to obtain the embedding $\mathbf{c}_{i}$, then $\mathbf{c}_{i}$ is treated as a condition and injected into the diffusion model. For a pre-trained SD model, $\mathbf{c}_{i}$ is further combined with a relevant text prompt $\mathcal{P}_{y_{i}}$ = `a photo of a $\mathcal{S}_{y_{i}}$' and sent to the UNet decoder to produce diffusion representations $\mathcal{F}_{i}$. Note that since we do not add noise, we simply set the time-steps $T=0$ of the diffusion process. 
Typically, the image $x_{i}$ size is 224$\times$224, and $\mathcal{F}_{i}$ contains 4 feature maps, where the $L$-th layer's feature map $\mathcal{F}_{i(L)}$ has the spatial size $\{7,14,28,28\}$ ($L \in \{1,2,3,4\}$), and $\{1280,1280,640,320\}$ are channels. Formally, conditional diffusion representations $\mathcal{F}_{i}$ are computed by the $\epsilon_{\theta}$ in Eq.~\eqref{eq:eq5}, image $x_{i}$, condition $\mathbf{c}_{i}$, text $\mathcal{P}_{y_{i}}$:
\begin{equation} \label{eq:eq7}
    \mathcal{F}_{i} = \epsilon_{\theta} (x_{i}, \tau_{i}, T=0, \mathbf{c}_{i}), \; \tau_{i} \gets \boldsymbol{\varphi} (\mathcal{P}_{y_{i}}),
\end{equation}
where $\boldsymbol{\varphi}$ denotes the text encoder of the SD model and $\tau_{i}$ is the text features with the prompt $\mathcal{P}_{y_{i}}$. 
Inspired by the results in Fig.~\ref{fig:fig3}, which motivate us to \textit{combine features at different layers of $\mathcal{F}_{i}$ to capture both semantics and details}. A simple concatenation, however, can lead to an unnecessarily high-dimensional feature ($1280+640+320=2240$). 
To reduce high dimension, we apply PCA for each layer $L \in \{2,3,4\}$ (due to the spatial resolution too low, we do not consider the $L=1$), and we aggregate them to the same resolution, then we connect them together to obtain the fused features $\widetilde{\mathcal{F}}_{i}$:
\begin{equation} \label{eq:eq8}
    \widetilde{\mathcal{F}}_{i} = \texttt{Concat}(\widetilde{\mathcal{F}}_{i(L)}), \; \widetilde{\mathcal{F}}_{i(L)} = \text{PCA}(\mathcal{F}_{i(L)}),
\end{equation}
where $L \in \{2,3,4\}$, we perform PCA: \romannumeral1) for $\mathcal{F}_{i(2)}$, we first upsample to the same resolution and apply PCA $\mathbb{R}^{1280\to256}$; \romannumeral2) for $\mathcal{F}_{i(3)}$ and $\mathcal{F}_{i(4)}$, we apply PCA $\mathbb{R}^{640 \to 128}$ and PCA $\mathbb{R}^{320 \to 128}$. 
Then, we combine them to form the final fused features $\widetilde{\mathcal{F}}_{i}$. 
As shown in the last column of Fig.~\ref{fig:fig4}, \textit{the fused features strike a balance between high-level and low-level visual concepts, focusing on both semantics and textures in the image}. On the example in the second row of Fig.~\ref{fig:fig4}, the fused feature yields more accurate matches (\textit{e.g.}, face and computer borders). The comparison results in Table.~\ref{tab:tab2} also confirm this viewpoint. 
We define $\widetilde{\mathcal{F}}_{i}$ calculated by Eq.~\eqref{eq:eq8} as \textit{conditional diffusion representations} of the sample $x_{i}$.

\subsection{Denoising Diffusion Representations} \label{sec:sec34}
Although we construct the conditional diffusion representations from the diffusion model's generative feedback, $\widetilde{\mathcal{F}}_{i}$ is still highly variable and depends on the condition generated by the local model, which can not ensure a consistent stable convergence signal. 
Based on Eq.~\eqref{eq:eq4}, we explore the inherent denoising property in diffusion models: SD model has learned extensive general knowledge about identical semantics via large pre-training, reflected in the UNet denoising process. 
Therefore, we aim to extract the denoising diffusion representations contained in a pre-trained SD model, to align the sample's embedding with the class semantics uniformly in the feature space. 
Given the image pair $\{x_{i}, y_{i}\}$ of $k$-th client's dataset $\mathcal{D}_{k}$, we first perform a forward process to add $t$ time-steps random Gaussian noise on $x_{i}$ as $\tilde{x}_{i}^{t} = \sqrt{\bar{\alpha}_{t}} x_{i} + \sqrt{1-\bar{\alpha}_{t}} \epsilon_{t}$, similar to~\cite{rombach2022high}, where $\bar{\alpha}_{t}$ and $\epsilon$ are the same as defined in the Eq.~\eqref{eq:eq3}. Next, we set the text prompt of $\tilde{x}_{i}^{t}$ as $\mathcal{P}_{y_{i}}$ = `a photo of a $\mathcal{S}_{y_{i}}$' used in the SD model pre-training~\cite{rombach2022high}, to avoid destroying correspondence between visual and linguistic concepts, and the $\mathcal{S}_{y_{i}}$ denotes the class name of the label $y_{i}$. 
Formally, the denoising diffusion representations $\widetilde{\mathcal{H}}_{i}$ of the sample $x_{i}$ are calculated by the $\epsilon_{\theta}$ in Eq.~\eqref{eq:eq4}, noisy sample $\tilde{x}_{i}^{t}$, text $\mathcal{P}_{y_{i}}$:
\begin{equation} \label{eq:eq9}
\begin{aligned}
    & \mathcal{H}_{i} = \epsilon_{\theta} (\tilde{x}_{i}^{t}, \tau_{i}, T=t), \; \tau_{i} \gets \boldsymbol{\varphi} (\mathcal{P}_{y_{i}}), \\
    & \widetilde{\mathcal{H}}_{i} = \texttt{Concat}(\widetilde{\mathcal{H}}_{i(L)}), \; \widetilde{\mathcal{H}}_{i(L)} = \text{PCA}(\mathcal{H}_{i(L)}),
\end{aligned}
\end{equation}
where $\boldsymbol{\varphi}$ denotes the text encoder of the SD model and $\tau_{i}$ is the text features of the text prompt $\mathcal{P}_{y_{i}}$. 
We further analyze the effects of the time-steps $t$ in Fig.~\ref{fig:fig6}. Then, we combine the features of layers $L \in \{2,3,4\}$, like Eq.~\eqref{eq:eq8}, to obtain fused features $\widetilde{\mathcal{H}}_{i}$. 
Intuitively, $\widetilde{\mathcal{H}}_{i}$ is considered as a virtual teacher to calibrate the embedding of $x_{i}$ at the feature-level. We define the denoising features $\widetilde{\mathcal{H}}_{i}$ calculated by Eq.~\eqref{eq:eq9} as \textit{denoising diffusion representations} of the sample $x_{i}$.
\section{Methodology}
\label{sec:method}

\begin{figure}[tbp]
    \centering
    \includegraphics[width=1.0\linewidth]{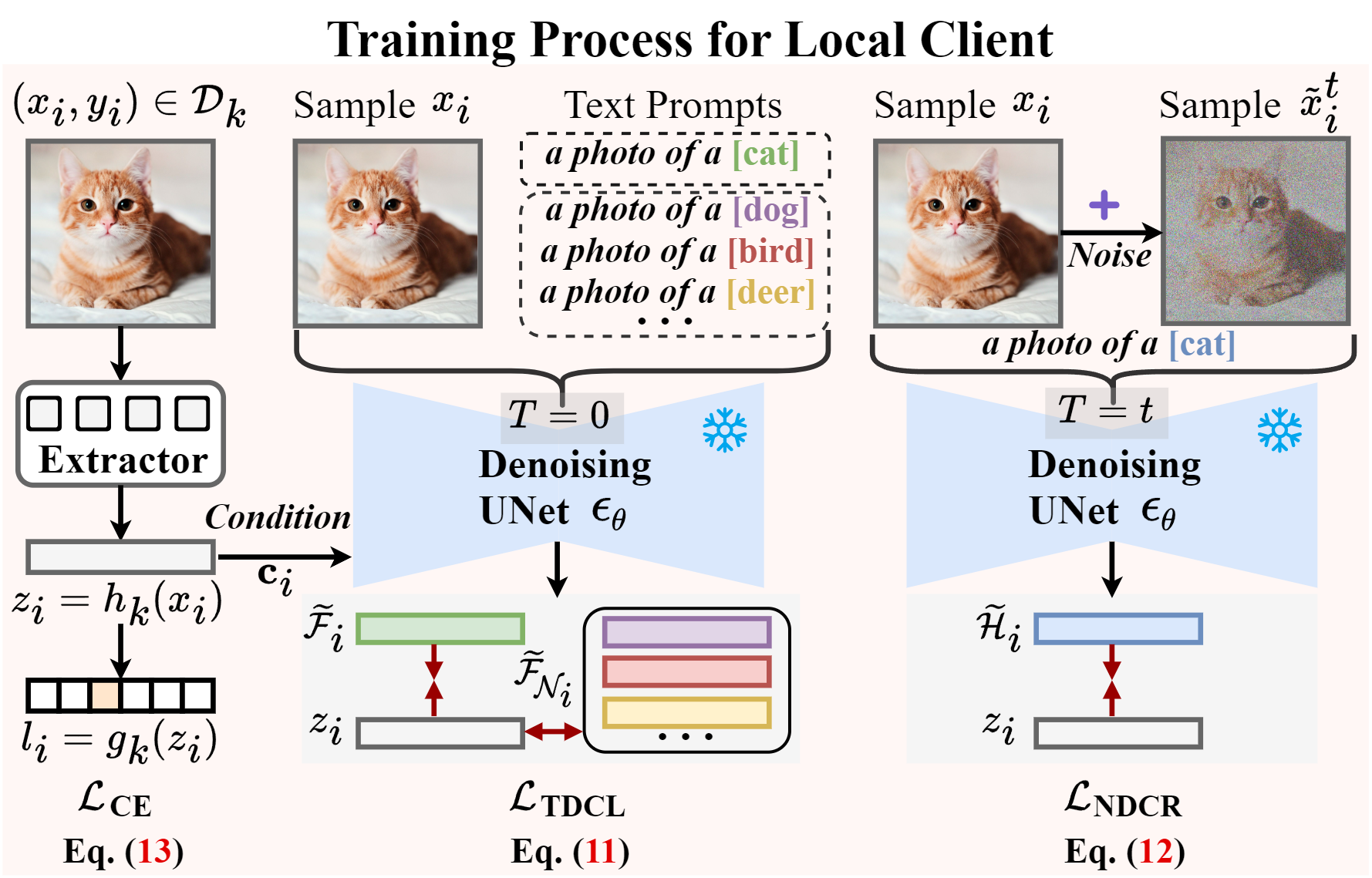}
    \caption{\textbf{Illustration of \texttt{FedDifRC}}, including two complementary modules: TDCL (Sec.~\ref{sec:tdcl}) builds an inter-contrastive learning strategy by conditional diffusion representations (Sec.~\ref{sec:sec33}), NDCR (Sec.~\ref{sec:ndcr}) designs a consistent regularization penalty via denoising diffusion representations (Sec.~\ref{sec:sec34}), jointly promoting FL training.}
    \label{fig:fig5}
\end{figure}

Our proposed \texttt{FedDifRC} leverage conditional and denoising diffusion representations to obtain diverse class-relevant knowledge and consistent convergence signals by two complementary components: Text-driven Diffusion Contrastive Learning in Sec.~\ref{sec:tdcl} and Noise-driven Diffusion Consistent Restriction in Sec.~\ref{sec:ndcr}. Further, we discuss how to extend \texttt{FedDifRC} to a self-supervised scheme (Sec.~\ref{sec:ssl}). Finally, we provide a theoretical analysis of \texttt{FedDifRC} under non-convex case (Sec.~\ref{sec:the}). Illustration of \texttt{FedDifRC} in Fig.~\ref{fig:fig5}.

\subsection{Text-Driven Diffusion Contrasting} \label{sec:tdcl}
Inspired by success of contrastive learning~\cite{chen2020simple,he2020momentum,li2021model}, based on Sec.~\ref{sec:sec33}, conditional diffusion representations $\widetilde{\mathcal{F}}_{i}$ of sample $\{x_{i}, y_{i}\} \in \mathcal{D}_{k}$ can be obtained with the condition $\mathbf{c}_{i}$ and the text prompt $\mathcal{P}_{y_{i}}$ = `a photo of a $\mathcal{S}_{y_{i}}$', where $\mathcal{S}_{y_{i}}$ is the class name of the label $y_{i}$. 
For each sample's embedding $z_{i}=h_{k}(x_{i}) \in \mathbb{R}^{d}$, we enforce $z_{i}$ to be similar to respective positive pair $\widetilde{\mathcal{F}}_{i}$ and dissimilar to negative sample pairs $\widetilde{\mathcal{F}}_{\mathcal{N}_{i}}$, where $\widetilde{\mathcal{F}}_{\mathcal{N}_{i}}$ means the conditional diffusion representations with the same condition $\mathbf{c}_{i}$ and the unmatched text prompts $\mathcal{P}_{j}$ = `a photo of a $\mathcal{S}_{j}$' ($\mathcal{S}_{j} \in \mathcal{S}$ and $\mathcal{S}_{j} \ne \mathcal{S}_{y_{i}}$). Then, we define the similarity between $z_{i}$ and $\widetilde{\mathcal{F}}_{i}$ as follows:
\begin{equation} \label{eq:eq10}
    s(z_{i}, \widetilde{\mathcal{F}}_{i})=\frac{z_{i} \cdot \widetilde{\mathcal{F}}_{i}}{\mathcal{U}(\|z_{i}\|_{2} \times \|\widetilde{\mathcal{F}}_{i}\|_{2})}, 
    \; \mathcal{U}=\frac{1}{n_{k}}\sum_{m=1}^{n_{k}} \|z_{m}-\widetilde{\mathcal{F}}_{i} \|_{2},
\end{equation}
where $n_{k}$ is total number of samples at client $k$, $\mathcal{U}$ is a normalization factor for similarity, where we set $\mathcal{U}$ to denote the average distance between $z_{m}|_{m=1}^{n_{k}}$ and $\widetilde{\mathcal{F}}_{i}$ on dataset $\mathcal{D}_{k}$. 
Next, we introduce TDCL to quantify and regulate the local model training, which can be formulated as follows:
\begin{equation} \label{eq:eq11}
    \mathcal{L}_{TDCL} = \log (1 + \frac{\sum_{\widetilde{\mathcal{F}}_{j} \in \widetilde{\mathcal{F}}_{\mathcal{N}_{i}}} \exp(s(z_{i}, \widetilde{\mathcal{F}}_{j})/\tau)}{\exp(s(z_{i}, \widetilde{\mathcal{F}}_{i})/\tau)}),
    % & \equiv \log (\sum_{\widetilde{\mathcal{F}}_{j} \in \widetilde{\mathcal{F}}_{\mathcal{N}_{i}}} \exp(s(z_{i}, \widetilde{\mathcal{F}}_{j})/\tau)) - \log (\exp(s(z_{i}, \widetilde{\mathcal{F}}_{i})/\tau)),
\end{equation}
where $\tau$ denotes a temperature parameter to control representation strength~\cite{chen2020simple}. By minimizing Eq.~\eqref{eq:eq11}, for client $k$, the local model brings each sample's embedding $z_{i}$ closer to the positive pair and away from other negative pairs.
%both generalizability and discriminability of the latent space during local training. TDCL adaptively enhances class-wise decision boundaries with distinct semantics in a condition-guided manner, promoting satisfying performance in FL.

\subsection{Noise-Driven Diffusion Regularization} \label{sec:ndcr}
Despite each local model aligning sample's embedding with conditional diffusion representations by optimizing $\mathcal{L}_{TDCL}$ in Eq.~\eqref{eq:eq11}. Since conditional diffusion representations are dynamically built from text-driven cross-modal matching and depend on generated condition $\mathbf{c}_{i}$ at each round, which could not offer a stable convergence direction. Therefore, we aim to leverage denoising diffusion representations ($\widetilde{\mathcal{H}}_{i}$ in Eq.~\eqref{eq:eq9}) to devise a consistent regularization penalty, constraining model's optimization region in the feature space. We design Noise-driven Diffusion Consistent Restriction (NDCR) and utilize a regularization term to pull the sample's embedding $z_{i}$ closer to the respective $\widetilde{\mathcal{H}}_{i}$:
\begin{equation} \label{eq:eq12}
    \mathcal{L}_{NDCR} = \sum_{q=1}^{d} (z_{i(q)}-\widetilde{\mathcal{H}}_{i(q)})^{2},
\end{equation}
where $q$ indexes the dimension $d$ of the embedding vector. Based on Eq.~\eqref{eq:eq12}, we expect to align the sample's embedding with the corresponding denoising diffusion representations, enforcing semantics uniformity of the feature space.

Besides, we construct CrossEntropy loss and use the logits output vector $l_{i}=g(z_{i})=f(x_{i})$ with the original label $y_{i}$ to optimize the local discriminative ability for each client. 
Formally, given the sample pair $\{x_{i}, y_{i}\}$ to calculate $\mathcal{L}_{CE}$:
\begin{equation} \label{eq:eq13}
    \mathcal{L}_{CE} = -\mathbf{1}_{y_{i}} \log (\psi(f(x_{i}))),
\end{equation}
where $\psi$ denotes the Softmax function, $f$ is the local model with parameters $\{u,v\}$ maps $x_{i}$ into a $|\mathcal{C}|$-dim output logits $l_{i}=f(x_{i})=g(h(x_{i}))$. 
Finally, for the $k$-th client, we carry out the following optimization objective in local training:
\begin{equation} \label{eq:eq14}
    \mathcal{L} = \mathcal{L}_{TDCL}(\text{Eq.~\eqref{eq:eq11}})+\mathcal{L}_{NDCR}(\text{Eq.~\eqref{eq:eq12}})+\mathcal{L}_{CE}.
\end{equation}
The algorithm of \texttt{FedDifRC} is given in Appendix \textcolor{refred}{A}.
%In each communication round, the local model is trained on the private dataset by optimizing $\mathcal{L}$ in Eq.~\eqref{eq:eq14}, then server collects all local weights to obtain the global weight, which is sent to each participant in the next round. The algorithm of \texttt{FedDifRC} (overview in Fig.~\ref{fig:fig5}) is given in Appendix \textcolor{refred}{A}.

\subsection{Expanding to Self-Supervised Scheme} \label{sec:ssl}
\texttt{FedDifRC} can be extended to the self-supervised scheme without relying on any labeled data source. 
Specifically, for $\mathcal{L}_{TDCL}$ in Eq.~\eqref{eq:eq11}, we only need to modify text template slightly: \romannumeral1) we set the text prompt $\mathcal{P}_{y_{i}}$ = `a photo of a similar object' in Eq.~\eqref{eq:eq7} to generate $\widetilde{\mathcal{F}}_{i}$ as the positive pair; \romannumeral2) then we randomly select from the set of all class names in Tiny-ImageNet~\cite{le2015tiny} to get $\mathcal{S}_{\mathcal{N}_{i}}$, and set $\mathcal{P}_{y_{i}}$ = `a photo of a $\mathcal{S}_{\mathcal{N}_{i}}$' to obtain $\widetilde{\mathcal{F}}_{\mathcal{N}_{i}}$ as negative pairs. For the $\mathcal{L}_{NDCR}$ in Eq.~\eqref{eq:eq12}, we set $\mathcal{P}_{y_{i}}$ = `a photo of a visual object' in Eq.~\eqref{eq:eq9} to get $\widetilde{\mathcal{H}}_{i}$ as consistent signal to align the sample's embedding. 
Note that it is also possible to use class names of other datasets (\textit{e.g.}, CIFAR100~\cite{cifar2009krizhevsky}) to generate $\widetilde{\mathcal{F}}_{\mathcal{N}_{i}}$ as negative pairs. 
By this self-supervised way, the objective term of local training becomes $\mathcal{L}_{TDCL}+\mathcal{L}_{NDCR}$. The results in Table.~\ref{tab:tab5} validate the effectiveness of \texttt{FedDifRC}'s self-supervised scheme.

\subsection{Convergence Analysis} \label{sec:the}
We proved convergence of \texttt{FedDifRC} in non-convex case. We denote $\mathcal{L}$ (Eq.~\eqref{eq:eq14}) as $\mathcal{L}_{r}$ with the round $r$, assumptions (Appendix \textcolor{refred}{B.2}) are similar to existing works~\cite{Li2020On,li2020fedprox,gao2022feddc,ye2023feddisco}.

% \textit{Convergence Rate of \texttt{FedDifRC}}: with the assumptions in Appendix \textcolor{refred}{B.2}, the round $r$ from $0$ to $R-1$, given any $\xi>0$, the objective function $\mathcal{L}$ will converge when,
\textit{Convergence Rate of \texttt{FedDifRC}}: Let the round $r$ from $0$ to $R-1$, given any $\xi>0$, the $\mathcal{L}$ will converge when,
\begin{equation} \label{eq:eq15}
    R > \frac{2(\mathcal{L}_{0}-\mathcal{L}^{*})}{\xi E \eta (2-L_{1} \eta) - \varOmega_{1} - \varOmega_{2}},
\end{equation}
and we can further get the following condition for $\eta$,
\begin{equation} \label{eq:eq16}
    \eta < \frac{2 \xi - 2(|\mathcal{C}|-1)L_{2}B}{L_{1}(\xi + \sigma^{2})},
\end{equation}
where $\varOmega_{1}=2(|\mathcal{C}|-1)L_{2}E \eta B$, $\varOmega_{2}=L_{1}E\eta^{2}\sigma^{2}$, and $\mathcal{L}^{*}$ denotes the optimal solution of $\mathcal{L}$, $\{L_{1},L_{2},B,\sigma^{2}\}$ are constants in the assumptions from Appendix \textcolor{refred}{B.2}, and $|\mathcal{C}|$ as the number of classes, $E$ as local epochs. 
Eq.~\eqref{eq:eq15} and Eq.~\eqref{eq:eq16} ensure convergence rate of \texttt{FedDifRC}, after selecting the number of rounds $R$ and the learning rate $\eta$. So the smaller $\xi$ is, the larger $R$ is, which means that the tighter the bound is, the more rounds $R$ is required. \textit{The detailed assumptions and proofs are formally provided in Appendix \textcolor{refred}{B}}.
\section{Experiments}

\begin{table}[t]
\renewcommand{\arraystretch}{1.0}
\centering
\resizebox{\linewidth}{!}{  % 根据宽高比自适应缩放
\begin{tabular}{c c | c c c c | c}
    \Xhline{1px}
    \rowcolor{mygray} &  & \multicolumn{5}{c}{\textbf{CIFAR10} (keep $\rho=1.0$)} \\
    \cline{3-7}
    \rowcolor{mygray} \multirow{-1.8}{*}{TDCL} & \multirow{-1.8}{*}{NDCR} 
        & $\mathtt{NID1}_{0.05}$ & $\mathtt{NID1}_{0.2}$ & $\mathtt{NID1}_{0.5}$ & $\mathtt{NID2}$ & AVG \\
    \hline
              &             & 78.27 & 84.65 & 86.11 & 72.60 & 80.41 \\
    \ding{51} &             & 81.39 & 86.03 & 88.16 & 75.67 & 82.81 \\
              & \ding{51}   & 80.35 & 86.40 & 87.54 & 75.33 & 82.40 \\
    \rowcolor{myye} \ding{51} & \ding{51} & 83.14 & 88.27 & 89.31 & 76.45 & \textbf{84.29} \\
    \hline\hline
    \rowcolor{mygray} &  & \multicolumn{5}{c}{\textbf{CIFAR10-LT} (keep $\mathtt{NID1}_{0.2}$)} \\
    \cline{3-7}
    \rowcolor{mygray} \multirow{-1.8}{*}{TDCL} & \multirow{-1.8}{*}{NDCR}
        & $\rho=10$ & $\rho=50$ & $\rho=100$ & $\rho=200$ & AVG \\
    \hline\hline
              &  & 75.94 & 63.25 & 52.97 & 49.69 & 60.46\\
    \ding{51} &  & 76.62 & 64.66 & 53.62 & 50.33 & 61.31\\
              & \ding{51} & 76.19 & 64.75 & 54.39 & 50.67 & 61.50\\
    \rowcolor{myye} \ding{51} & \ding{51} & 77.58 & 65.30 & 55.01 & 51.12 & \textbf{62.25}\\
    \Xhline{1px}
\end{tabular}}
% \vspace{-3.36mm}
\caption{Ablation study for two key modules: TDCL and NDCR.}
\label{tab:tab1}
% \vspace{-3.36mm}
\end{table}

\begin{table}[t]
\renewcommand{\arraystretch}{1.0}
\centering
\resizebox{\linewidth}{!}{  % 根据宽高比自适应缩放
\begin{tabular}{l | c c c c | c c}
    \Xhline{1px}
    \rowcolor{mygray} Scenarios & $L=1$ & $L=2$ & $L=3$ & $L=4$ & \textit{Fused} & $\triangle$ \\
    \hline
    $\mathtt{NID1}_{0.05}$ & 80.21 & \underline{82.60} & 82.19 & 81.33 & \cellcolor{myye}{\textbf{83.14}} & $+$ 0.54 \\
    $\mathtt{NID1}_{0.2}$  & 86.15 & 87.28 & \underline{87.81} & 85.92 & \cellcolor{myye}{\textbf{88.27}} & $+$ 0.46 \\
    $\mathtt{NID1}_{0.5}$  & 87.10 & \underline{89.03} & 88.93 & 87.17 & \cellcolor{myye}{\textbf{89.31}} & $+$ 0.28 \\
    $\mathtt{NID2}$  & 74.26 & \underline{75.73} & 75.61 & 75.00 & \cellcolor{myye}{\textbf{76.45}} & $+$ 0.72 \\
    \Xhline{1px}
\end{tabular}}
% \vspace{-3.36mm}
\caption{Comparison of features from different layers $L$ and fused features ($L \in \{2,3,4\}$) in Eq.~\eqref{eq:eq8} and Eq.~\eqref{eq:eq9} to obtain $\widetilde{\mathcal{F}}_{i}$ and $\widetilde{\mathcal{H}}_{i}$, \textbf{best} in bold and \underline{second} with underline, $\triangle$ means accuracy gap.}
\label{tab:tab2}
% \vspace{-3.36mm}
\end{table}

\begin{figure}[t]
\centering
    \begin{minipage}[t]{0.495\linewidth}
        \centering
        \includegraphics[width=1.0\linewidth]{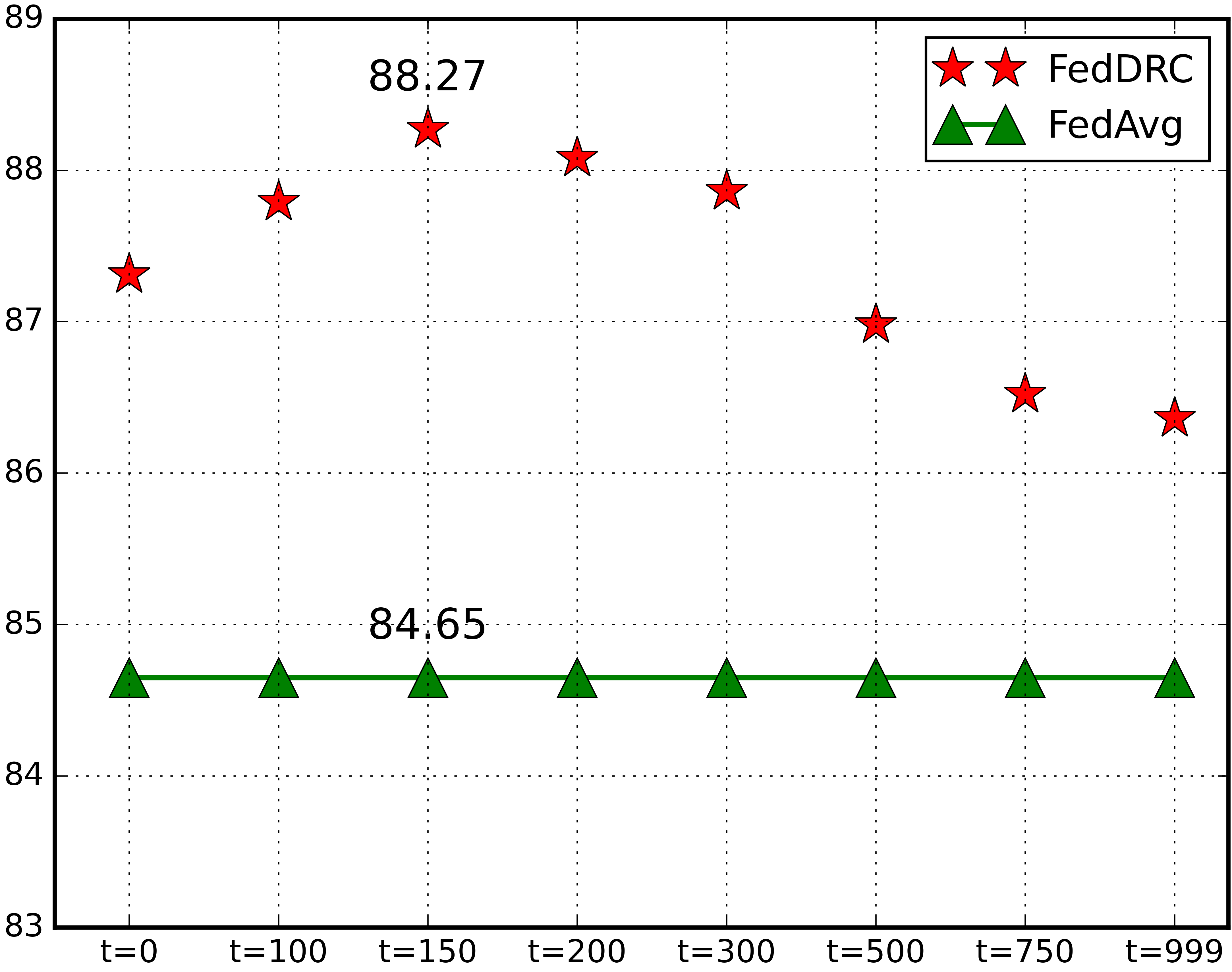}
    \end{minipage}
    \begin{minipage}[t]{0.495\linewidth}
        \centering
        \includegraphics[width=1.0\linewidth]{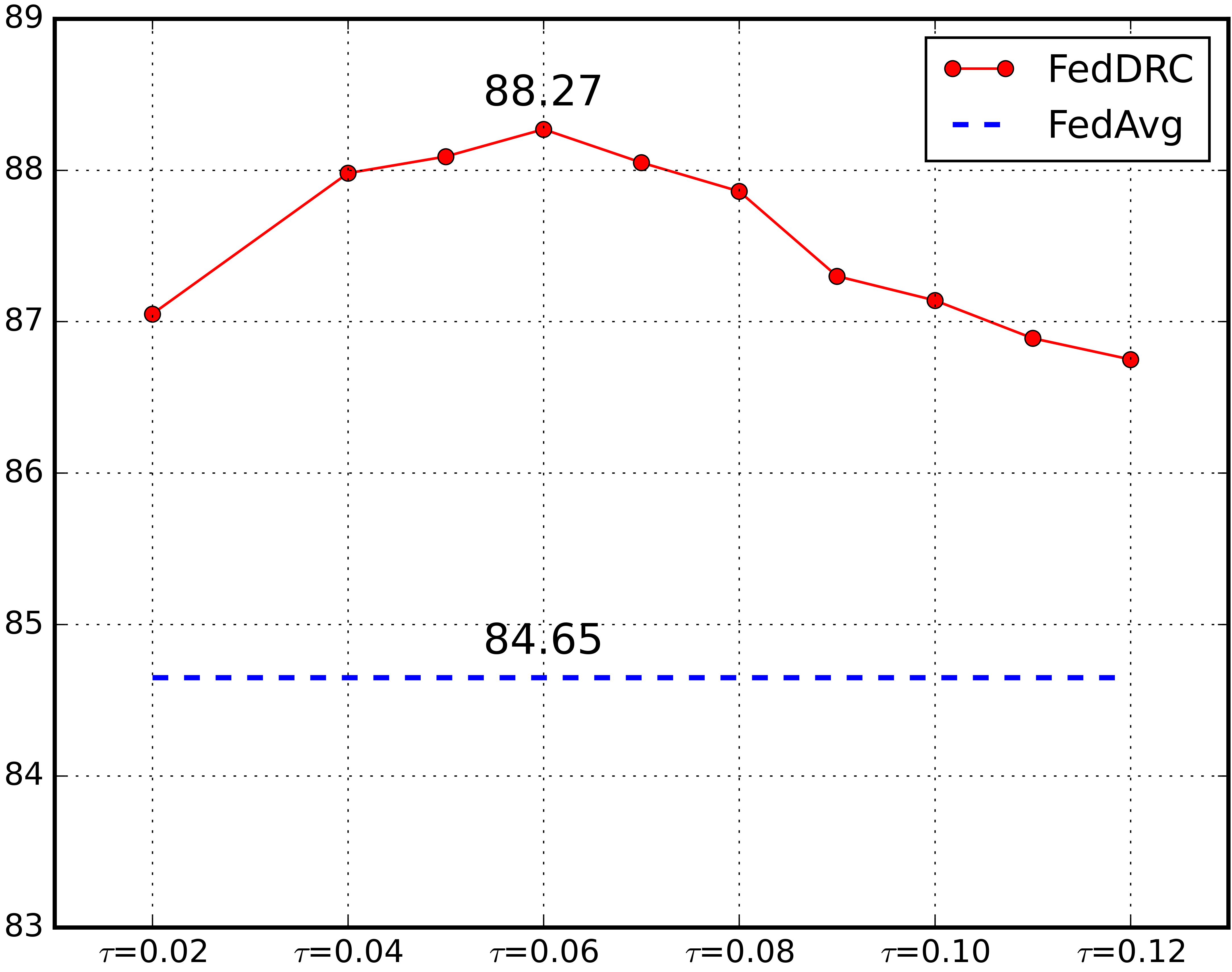}
    \end{minipage}
\caption{Analysis of \texttt{FedDifRC} with time-steps $t$ in Eq.~\eqref{eq:eq9} (\textbf{left}) and temperature $\tau$ in Eq.~\eqref{eq:eq11} (\textbf{right}) on CIFAR10 with $\mathtt{NID1}_{0.2}$.}
% \vspace{-2.8mm}
\label{fig:fig6}
\end{figure}

\begin{table*}[t]
\renewcommand{\arraystretch}{1.0}
\centering
\resizebox{\linewidth}{!}{  % 根据宽高比自适应缩放
\begin{tabular}{lccccccccccccc}
    \toprule[1.0pt]
    \rowcolor{mygray} & \multicolumn{3}{c}{\textbf{CIFAR10}} & \multicolumn{3}{c}{\textbf{CIFAR100}} & \multicolumn{3}{c}{\textbf{TinyImageNet}} & \multicolumn{4}{c}{\textbf{CIFAR10-LT} (keep $\mathtt{NID1}_{0.2}$)} \\
    \cmidrule(lr){2-4} \cmidrule(lr){5-7} \cmidrule(lr){8-10} \cmidrule(lr){11-14}
    \rowcolor{mygray} \multirow{-3.0}{*}{\textbf{FL Baselines}} & $\mathtt{NID1}_{0.05}$ & $\mathtt{NID1}_{0.2}$ & $\mathtt{NID2}$ & $\mathtt{NID1}_{0.05}$ & $\mathtt{NID1}_{0.2}$ & $\mathtt{NID2}$ & $\mathtt{NID1}_{0.05}$ & $\mathtt{NID1}_{0.2}$ & $\mathtt{NID2}$ & $\rho=10$ & $\rho=50$ & $\rho=100$ & $\rho=200$ \\
    \cmidrule(lr){1-1} \cmidrule(lr){2-4} \cmidrule(lr){5-7} \cmidrule(lr){8-10} \cmidrule(lr){11-14}
    $\text{FedAvg}_{[\text{AISTAT'17}]}$ & 78.27 & 84.65 & 72.60 & 55.97 & 60.08 & 50.56 & 40.41 & 42.84 & 35.15 & 75.94 & 63.25 & 52.97 & 49.69 \\
    $\text{FedProx}_{[\text{MLSys'20}]}$ & 78.42 & 84.59 & 72.81 & 56.27 & 60.21 & 50.29 & 40.20 & 42.16 & 35.62 & 76.02 & 63.43 & 52.86 & 49.10 \\
    $\text{MOON}_{[\text{CVPR'21}]}$ & 80.79 & 86.10 & 73.35 & 56.79 & 61.48 & 51.81 & 40.79 & 43.63 & 36.11 & 75.73 & 63.80 & 53.17 & 49.31 \\
    $\text{FedProto}_{[\text{AAAI'22}]}$ & 77.86 & 83.90 & 70.84 & 54.48 & 59.26 & 48.60 & 39.29 & 41.02 & 34.68 & 74.31 & 62.23 & 50.37 & 47.26 \\
    $\text{FedNH}_{[\text{AAAI'23}]}$ & 80.25 & 85.82 & 73.65 & 56.68 & 61.45 & 51.31 & 40.25 & 43.30 & 36.21 & 76.15 & 63.36 & 52.49 & 48.92 \\
    $\text{FedDisco}_{[\text{ICML'23}]}$ & 81.35 & 86.63 & 74.72 & 57.34 & 61.79 & 52.05 & 40.83 & 43.92 & 37.69 & 76.27 & \underline{64.40} & 53.33 & 49.75 \\
    $\text{FedCDA}_{[\text{ICLR'24}]}$ & 81.70 & 86.89 & 75.07 & 57.76 & \underline{62.17} & 52.35 & 41.18 & 44.09 & 38.24 & \underline{76.45} & 64.30 & \underline{53.85} & \underline{50.08} \\
    $\text{FedRCL}_{[\text{CVPR'24}]}$ & \underline{82.02} & \underline{87.11} & \underline{75.13} & \underline{57.98} & 62.05 & \underline{52.49} & \underline{41.33} & \underline{44.21} & \underline{38.86} & 76.34 & 64.17 & 53.41 & 49.91 \\
    \cmidrule(lr){1-14}
    \rowcolor{myye} \texttt{FedDifRC} & \textbf{83.14} & \textbf{88.27} & \textbf{76.45} & \textbf{59.22} & \textbf{63.32} & \textbf{53.87} & \textbf{42.63} & \textbf{45.28} & \textbf{40.05} & \textbf{77.58} & \textbf{65.30} & \textbf{55.01} & \textbf{51.12} \\
    \bottomrule[1.0pt]
\end{tabular}}
% \vspace{-3mm}
\caption{Comparison results on several settings (label shift and imbalance shift) and datasets, the \textbf{best} in bold and \underline{second} with underline.}
\label{tab:tab3}
% \vspace{-4mm}
\end{table*}

\begin{table}[t]
\renewcommand{\arraystretch}{1.0}
\centering
\resizebox{\linewidth}{!}{  % 根据宽高比自适应缩放
\begin{tabular}{lccccc}
    \toprule[1.0pt]
    \rowcolor{mygray} & \multicolumn{5}{c}{\textbf{Office-Caltech}} \\
    \cmidrule(lr){2-6}
    \rowcolor{mygray} \multirow{-3.0}{*}{\textbf{FL Baselines}} & Caltech & Webcam & Amazon & DSLR & AVG \\
    \cmidrule(lr){1-1} \cmidrule(lr){2-5} \cmidrule(lr){6-6}
    $\text{FedDisco}_{[\text{ICML'23}]}$ & 62.39 & 45.95 & 77.31 & 45.20 & 57.71 \\
    $\text{FedCDA}_{[\text{ICLR'24}]}$ & 62.25 & 46.53 & 76.58 & 46.14 & \underline{57.87} \\
    $\text{FedRCL}_{[\text{CVPR'24}]}$ & 60.13 & 42.76 & 75.60 & 43.77 & 55.56 \\
    \cmidrule(lr){1-6}
    \rowcolor{myye} \texttt{FedDifRC} & 63.32 & 50.19 & 78.67 & 48.12 & \textbf{60.08} \\
    \bottomrule[1.0pt]
\end{tabular}}
% \vspace{-3mm}
\caption{Comparison results on Office-Caltech with domain shift.}
\label{tab:tab4}
% \vspace{-3.36mm}
\end{table}

\begin{table}[t]
\renewcommand{\arraystretch}{1.0}
\centering
\resizebox{\linewidth}{!}{  % 根据宽高比自适应缩放
\begin{tabular}{lccccc}
    \toprule[1.0pt]
    \rowcolor{mygray} & & \multicolumn{4}{c}{\textbf{CIFAR100}} \\
    \cmidrule(lr){3-6}
    \rowcolor{mygray} \multirow{-3.0}{*}{\textbf{FL Baselines}} & \multirow{-3.0}{*}{\textbf{Clients}} & $\mathtt{NID1}_{0.05}$ & $\mathtt{NID1}_{0.2}$ & $\mathtt{NID1}_{0.5}$ & $\mathtt{NID2}$ \\
    \cmidrule(lr){1-1} \cmidrule(lr){2-2} \cmidrule(lr){3-6}
    $\text{FedEMA}_{[\text{ICLR'22}]}$ & $K=10$ & 26.90 & 29.41 & 30.64 & 21.52 \\
    $\text{FedLID}_{[\text{ICCV'23}]}$ & $K=10$ & 28.84 & 30.43 & 31.22 & 23.65 \\
    \cmidrule(lr){1-6}
    \rowcolor{myye} \texttt{FedDifRC} & $K=10$ & \textbf{30.61} & \textbf{31.45} & \textbf{32.30} & \textbf{25.73} \\
    \bottomrule[1.0pt]
\end{tabular}}
% \vspace{-3mm}
\caption{Comparison results on self-supervised setting (Sec.~\ref{sec:ssl}).}
\label{tab:tab5}
% \vspace{-3.36mm}
\end{table}

\begin{table}[t]
\renewcommand{\arraystretch}{1.0}
\centering
\resizebox{\linewidth}{!}{  % 根据宽高比自适应缩放
\begin{tabular}{lcccccc}
    \toprule[1.0pt]
    \rowcolor{mygray} & \multicolumn{3}{c}{\textbf{CIFAR10}} & \multicolumn{3}{c}{\textbf{CIFAR10-LT} ($\mathtt{NID1}_{0.2}$)} \\
    \cmidrule(lr){2-4} \cmidrule(lr){5-7}
    \rowcolor{mygray} \multirow{-3.0}{*}{\textbf{FL Baselines}} & $\mathtt{NID1}_{0.05}$ & $\mathtt{NID1}_{0.2}$ & $\mathtt{NID2}$ & $\rho=10$ & $\rho=50$ & $\rho=100$ \\
    \cmidrule(lr){1-1} \cmidrule(lr){2-4} \cmidrule(lr){5-7}
    $\text{FedDDA}_{[\text{PRCV'23}]}$ & 82.05 & 86.94 & 74.49 & 76.31 & 63.95 & 53.15 \\
    $\text{FedDISC}_{[\text{AAAI'24}]}$ & 82.33 & 87.15 & 75.42 & 76.60 & 64.23 & 54.07 \\
    \cmidrule(lr){1-7}
    FedRCL ($\diamondsuit$ / $\dagger$) & $\text{82.30}^{\diamondsuit}$/$\text{80.95}^{\dagger}$ & $\text{87.25}^{\diamondsuit}$/$\text{86.16}^{\dagger}$ & $\text{75.37}^{\diamondsuit}$/$\text{73.85}^{\dagger}$ & $\text{76.50}^{\diamondsuit}$/$\text{76.12}^{\dagger}$ & $\text{64.26}^{\diamondsuit}$/$\text{62.91}^{\dagger}$ & $\text{53.65}^{\diamondsuit}$/$\text{52.20}^{\dagger}$ \\
    FedRCL ($\diamondsuit$ + $\dagger$) & 81.52 & 86.48 & 74.35 & 76.22 & 63.69 & 52.93 \\
    \cmidrule(lr){1-7}
    \rowcolor{myye} \texttt{FedDifRC} & \textbf{83.14} & \textbf{88.27} & \textbf{76.45} & \textbf{77.58} & \textbf{65.30} & \textbf{55.01} \\
    \bottomrule[1.0pt]
\end{tabular}}
% \vspace{-3mm}
\caption{More comparison results on CIFAR10 and CIFAR10-LT.}
\label{tab:tab6}
% \vspace{-3.36mm}
\end{table}

\begin{table}[t]
\renewcommand{\arraystretch}{1.0}
\centering
\resizebox{\linewidth}{!}{  % 根据宽高比自适应缩放
\begin{tabular}{lcccccc}
    \toprule[1.0pt]
    \rowcolor{mygray} & \multicolumn{3}{c}{\textbf{HAM10000}~\cite{tschandl2018ham10000}} & \multicolumn{3}{c}{\textbf{EuroSAT}~\cite{helber2019eurosat}} \\
    \cmidrule(lr){2-4} \cmidrule(lr){5-7}
    \rowcolor{mygray} \multirow{-3.0}{*}{\textbf{FL Baselines}} & $\mathtt{NID1}_{0.2}$ & $\mathtt{NID1}_{0.5}$ & $\mathtt{NID2}$ & $\mathtt{NID1}_{0.2}$ & $\mathtt{NID1}_{0.5}$ & $\mathtt{NID2}$ \\
    \cmidrule(lr){1-1} \cmidrule(lr){2-4} \cmidrule(lr){5-7}
    $\text{FedDisco}_{[\text{ICML'23}]}$ & 50.26 & 53.68 & 44.63 & 78.76 & 82.36 & 70.34 \\
    $\text{FedCDA}_{[\text{ICLR'24}]}$ & 51.65 & 54.05 & 45.16 & \underline{80.45} & 83.38 & \underline{71.10} \\
    $\text{FedRCL}_{[\text{CVPR'24}]}$ & \underline{52.21} & \underline{54.20} & \underline{45.48} & 79.80 & \underline{83.69} & 70.73 \\
    \cmidrule(lr){1-7}
    \rowcolor{myye} \texttt{FedDifRC} & \textbf{53.79} & \textbf{55.60} & \textbf{47.17} & \textbf{82.49} & \textbf{86.05} & \textbf{73.14} \\
    \bottomrule[1.0pt]
\end{tabular}}
% \vspace{-3mm}
\caption{Comparison results on the HAM10000 and EuroSAT.}
\label{tab:tab7}
% \vspace{-3.36mm}
\end{table}

\begin{table}[t]
\renewcommand{\arraystretch}{1.0}
\centering
\resizebox{\linewidth}{!}{  % 根据宽高比自适应缩放
\begin{tabular}{lcccccc}
    \toprule[1.0pt]
    \rowcolor{mygray} & \multicolumn{3}{c}{\textbf{CIFAR10}} & \multicolumn{3}{c}{\textbf{CIFAR10-LT} ($\mathtt{NID1}_{0.2}$)} \\
    \cmidrule(lr){2-4} \cmidrule(lr){5-7}
    \rowcolor{mygray} \multirow{-3.0}{*}{\textbf{Pre-trained Models}} & $\mathtt{NID1}_{0.05}$ & $\mathtt{NID1}_{0.2}$ & $\mathtt{NID2}$ & $\rho=10$ & $\rho=50$ & $\rho=100$ \\
    \cmidrule(lr){1-1} \cmidrule(lr){2-4} \cmidrule(lr){5-7}
    CLIP-ViT-B/16~\cite{radford2021learning} & 79.72 & 84.91 & 73.97 & 74.89 & 62.44 & 50.87 \\
    MAE-ViT-B/16~\cite{he2022masked} & 80.74 & 85.38 & 73.24 & 75.45 & 63.27 & 52.55 \\
    DINO-ViT-B/16~\cite{caron2021emerging} & 80.57 & 86.14 & 74.16 & 75.25 & 62.70 & 51.92 \\
    \cmidrule(lr){1-7}
    \rowcolor{myye} \texttt{FedDifRC} & \textbf{83.14} & \textbf{88.27} & \textbf{76.45} & \textbf{77.58} & \textbf{65.30} & \textbf{55.01} \\
    \bottomrule[1.0pt]
\end{tabular}}
% \vspace{-3mm}
\caption{Experiments on aligning features of pre-trained models.}
\label{tab:tab8}
% \vspace{-3.36mm}
\end{table}

\subsection{Experimental Setup}
\textbf{Datasets:} We adopt popular benchmark datasets, including CIFAR10 \& CIFAR100~\cite{cifar2009krizhevsky}, TinyImageNet~\cite{le2015tiny}; we build unbalanced versions of these datasets by~\cite{caoKaidi2019}: CIFAR10-LT \& CIFAR100-LT, and TinyImageNet-LT; we use Digits~\cite{peng2019moment} and Office-Caltech~\cite{gong2012geodesic} for the domain shift task in FL. \\
\textbf{Scenarios:} \romannumeral1) \textit{label shift}: we consider two \emph{non-iid} settings: $\mathtt{NID1}_{\alpha}$ follows Dirichlet distribution~\cite{wang2020fedma}, where $\alpha$ denotes the heterogeneity level; $\mathtt{NID2}$ is a more extreme setting consists of 6 biased clients (each has only a single category) and 1 unbiased client has all categories. 
\romannumeral2) \textit{imbalance shift}: we shape the original dataset into a long-tailed distribution with $\rho$ follow~\cite{caoKaidi2019}, $\rho$ means the ratio between sample sizes of the most frequent and least frequent class, as $\rho=\max_{j} / \min_{j}$. 
\romannumeral3) \textit{domain shift}: Digits~\cite{peng2019moment} includes 4 domains as MNIST (M), USPS (U), SVHN (SV), SYN (SY) with 10 categories. Office-Caltech~\cite{gong2012geodesic} also consists 4 domains as Caltech (C), Webcam (W), Amazon (A), DSLR (D) with 10 overlapping categories. We initialize 10 clients and assign them domains randomly: Digits (M: 1, U: 4, SV: 3, SY: 2), Office-Caltech (C: 3, W: 2, A: 2, D: 3). Each client's dataset is randomly selected from domains (Digits: 10\%, Office-Caltech: 20\%). \\
\textbf{Configurations:} The global rounds $R$ as 100 and clients $K$ as 10, all methods have little or no gain with more rounds $R$. To facilitate fair comparison, we follow in~\cite{li2020fedprox,li2021model}, and use SGD optimizer to update models with a learning rate $0.01$, a momentum of $0.9$, a batch size of $64$, the weight decay of $10^{-5}$, and local epochs $E=10$. We use MobileNetV2~\cite{sandler2018mobilenetv2} for TinyImageNet, ResNet-10~\cite{resnet8ju} for others. The dimension $d$ as 512, $\tau$ (Eq.~\eqref{eq:eq11}) as 0.06, $t$ (Eq.~\eqref{eq:eq9}) as 150. \\
\textbf{Baselines}: We compare our against several SOTA methods: standard-based FedAvg~\cite{mcmahan2017communication}, FedProx~\cite{li2020fedprox}; contrast-based MOON~\cite{li2021model}, FedRCL~\cite{seo2024relaxed}; aggregate-based FedCDA~\cite{wang2023fedcda}, FedDisco~\cite{ye2023feddisco}; unsupervised FedEMA~\cite{zhuang2022divergence}, FedLID~\cite{psaltis2023fedlid}; prototype-based FedProto~\cite{tan2022fedproto}, FedNH~\cite{dai2023tackling}.

\subsection{Validation Analysis}
We give ablation study in Table.~\ref{tab:tab1}: \romannumeral1) TDCL brings significant performance improvements over the baseline, showing that TDCL is able to promote semantic spread-out property; \romannumeral2) NDCR also yields solid gains, proving the importance of aligning instance embedding via consistency regularization; \romannumeral3) combining TDCL and NDCR attain better performance, supporting our motivation to unleash meaningful guidance of diffusion representations to promote FL local training.

Then, we prove the advantage of fused features (Eq.~\eqref{eq:eq8}). As shown in Table.~\ref{tab:tab2}, results on CIFAR10 reveal that fused features capture both semantics and local details, better than features on a single layer $L$, and if the features with too low resolution ($L$=1) or over-matching ($L$=4) would damage the model's performance. To explore the influence of different time-steps $t$ (Eq.~\eqref{eq:eq9}), we conduct experiments on CIFAR10 with $\mathtt{NID1}_{0.2}$ in Fig.~\ref{fig:fig6} left, we add different amount of noise to get $\tilde{x}_{i}^{t}$ in Eq.~\eqref{eq:eq9}. We find that \texttt{FedDifRC} is robust to the choice of $t \in [100, 300]$ ($t$=150 achieves the highest value). 
As shown in Fig.~\ref{fig:fig6} right, we present results with different $\tau$ in Eq.~\eqref{eq:eq11}, which reveals that \texttt{FedDifRC} is not sensitive to the $\tau \in [0.04,0.08]$, indicating the effectiveness of $\mathcal{U}$ (Eq.~\eqref{eq:eq10}), to balance relative distances among different samples. The accuracy no longer increases after $\tau$=0.06. Extra results of $t$ and $\tau$ are provided in Appendix \textcolor{refred}{C.2.5}.
% For CIFAR100, \texttt{FedDifRC} achieves the optimal accuracy when $t$=150 and $\tau$=0.06. For TinyImageNet, the accuracy of $t$=200, $\tau$=0.06 is slightly better than $t$=150, $\tau$=0.06. To prevent the redundancy of adjusting parameter, we thus set $t$=150 and $\tau$=0.06 in our experiments as default values.

\subsection{Evaluation Results}
The results on different settings ($\mathtt{NID1}$, $\mathtt{NID2}$, $\rho$) in Table.~\ref{tab:tab3}, \texttt{FedDifRC} consistently outperforms these counterparts, it confirms that \texttt{FedDifRC} effectively leverages the inherent fine-grained visual representation capabilities of diffusion models to promote heterogeneous FL training. The results of Office-Caltech with domain shift are shown in Table.~\ref{tab:tab4}, indicating that \texttt{FedDifRC} can acquire well-generalizable property in the feature space. With self-supervised scenario, in Table.~\ref{tab:tab5}, our \texttt{FedDifRC} achieves excellent performance by leveraging the visual perception capability of diffusion models, without requiring any labeled data. 
Please refer to Appendix \textcolor{refred}{C} for more details and results (\textit{e.g.}, domain shift, convergence rates, different SD versions, etc).
% The results with the imbalance shift on CIFAR100-LT and TinyImageNet-LT datasets, and the results with domain shift on the Digits, are given in Appendix \textcolor{refred}{C.2.2} and \textcolor{refred}{C.2.3}. 
% We also compare the convergence rate of \texttt{FedDifRC} and baselines, shown in Appendix \textcolor{refred}{C.2.4}, indicating the superiority of \texttt{FedDifRC} in dealing with heterogeneous FL scenarios and the efficiency for FL model training. 
% The ablation study on the number of clients $K$ and local epochs $E$ for \texttt{FedDifRC} are provided in Appendix \textcolor{refred}{C.2.7} and \textcolor{refred}{C.2.8}. 
% Since our \texttt{FedDifRC} is built on the SD model, we also investigate how the SD model's weights would affect the performance of \texttt{FedDifRC}. In our previous experiments, SD-1.5 as basic SD model, we then compare different versions (\textit{e.g.}, SD-1.4, SD-2.0) of the SD model on CIFAR10, and the results are shown in Appendix \textcolor{refred}{C.2.9} (\textit{e.g.}, SD-1.4 get 83.09\% and 88.23\% of $\mathtt{NID1}_{0.05}$ and $\mathtt{NID1}_{0.2}$). Our initial purpose is not to generate the image, different SD model versions have almost little difference.
% Our initial purpose is not to generate the image, different versions of the SD model have almost little difference. \textit{Please see Appendix \textcolor{refred}{C} for more details and results}.

To further elaborate on the core idea of our \texttt{FedDifRC}, we then perform the following experiments:\\
\textit{\romannumeral1) To ensure a fair comparison}, we also compare diffusion-based FL methods FedDISC~\cite{yang2024exploring} and FedDDA~\cite{zhao2023federated}, we generate data for each client (same amount) by the SD model as $\diamondsuit$, and align SD features (the image is fed directly into the SD model) by L2 loss as $\dagger$. Table.~\ref{tab:tab6} shows that directly matching SD features may even hurt the performance, which further illustrates the superiority of our \texttt{FedDifRC}. \\
\textit{\romannumeral2) We focus on visual perception ability of diffusion models to understand an image (\textit{e.g.}, textures and structures) rather than pre-trained prior knowledge.} 
We conduct experiments on medical and remote sensing datasets HAM10000~\cite{tschandl2018ham10000} and EuroSAT~\cite{helber2019eurosat}, which are not crossed in SD pre-training dataset LAION-5B~\cite{schuhmann2022laion}. Table.~\ref{tab:tab7} show that the success of \texttt{FedDifRC} is based on the fine-grained visual representation rather than the large prior knowledge in SD model. \\
\textit{\romannumeral3) The SD model can implicitly provide both high-level and low-level visual concepts of semantic objects, while foundation models focus more on sparse visual matching.} 
The Table.~\ref{tab:tab8} show that \texttt{FedDifRC} can capture underlying properties and correspondences of visual objects.
\section{Conclusion}
\label{sec:conclusion}
In this paper, we aim to exploit the fine-grained visual representation capabilities of diffusion models to handle data heterogeneity in FL. 
We propose a novel diffusion-inspired FL framework \texttt{FedDifRC} to regulate local training via denoising and conditional diffusion representations. 
By leveraging complementary strength of TDCL and NDCR, \texttt{FedDifRC} achieves superior performance in heterogeneous FL.

\section*{Acknowledgment}
This work is supported by the scholarship from the China Scholarship Council (CSC) while the first author pursues his PhD degree at the University of Wollongong. This work was also partially supported by Australian Research Council Linkage Project LP210300009 and LP230100083.

{
    \small
    \bibliographystyle{ieeenat_fullname}
    \bibliography{main}
}

\end{document}